\documentclass[11pt,letterpaper,onecolumn]{article}
\usepackage[T1]{fontenc}
\usepackage[margin=1in]{geometry}
\usepackage{newtxtext,newtxmath}
\usepackage{graphicx}
\usepackage{booktabs,tabularx,array}
\usepackage{caption}
\usepackage[section]{placeins}
\usepackage[hidelinks]{hyperref}
\usepackage{microtype}

\title{Anatomy-Aware Synthesis of Post-Contrast Breast MRI from Pre-Contrast Images}
\author{Zhengbo Zhou$^{1,\star,\#}$, Dooman Arefan$^{1,\dagger,\#}$, Lin Gu$^{2}$, Ufara Zuwasti Curran$^{3}$, Shandong Wu$^{1,\star,\dagger,\mathsection}$}
\date{}

\makeatletter
\newcommand{\institute}[1]{\gdef\@institute{#1}}
\institute{%
$^{1}$ University of Pittsburgh, Pittsburgh, PA, USA\\
$^{\star}$ Intelligent Systems Program \quad \\
$^{\dagger}$ Department of Radiology \quad \\
$^{\mathsection}$ Department of Biomedical Informatics \& Bioengineering\\[3pt]
$^{2}$ Mallinckrodt Institute of Radiology,\\
Washington University School of Medicine, St. Louis, MO, USA\\[3pt]
$^{3}$ Fairfax Radiological Consultants, P.L.L.C
}

\renewcommand{\maketitle}{%
  \begin{center}
    \Large\bfseries\@title\par
    \ifx\@author\@empty\else
      \smallskip{\normalsize\normalfont\@author\par}
    \fi
    \smallskip{%
      \normalfont\fontsize{9}{10}\selectfont\@institute\par
    }
  \end{center}
  \begingroup
    \renewcommand{\thefootnote}{\#}
    \footnotetext[0]{%
      Zhengbo Zhou and Dooman Arefan contributed equally
      to this work.
    }
  \endgroup
}
\makeatother

\renewenvironment{abstract}{%
  \par\begingroup\small\noindent\textbf{Abstract}\par\medskip
}{\par\endgroup}

\hypersetup{%
  pdfauthor={Zhengbo Zhou, Dooman Arefan, Lin Gu, Ufara Zuwasti Curran, Shandong Wu},
  pdftitle={From Pre- to Post-Contrast Breast MRI: Anatomy-Aware Image-To-Image Translation}
}

\begin{document}
\begingroup
\setlength{\parskip}{0.15em}
\maketitle

\begin{abstract}
\noindent This study aimed to develop and evaluate an anatomy-aware deep learning framework for synthesizing post-contrast breast MRI from pre-contrast images, with the goal of improving fidelity in tumor and background parenchymal enhancement (BPE) regions. In this retrospective study, breast MRI data from 649 patients comprising 6,251 paired images with available tumor-area slices were used to train and evaluate a synthesis model that generates post-contrast images from pre-contrast inputs, with an original image size of 512 \(\times\) 512 pixels. The proposed method used an image-to-image translation framework by incorporating anatomy-aware constraints through breast mask consistency, lesion-region supervision, and BPE-region supervision. Quantitative evaluation included structural similarity index measure (SSIM), peak signal-to-noise ratio (PSNR), and normalized root mean square error (NRMSE), with comparisons against Pix2Pix, Pix2PixHD, diffusion-based synthesis, Pix2PixHD with tumor-mask supervision, and Pix2PixHD with tumor- and BPE-mask supervision. A reader study involving two breast radiologists assessed perceptual differences of the synthetic images. Downstream analyses examined whether synthetic postcontrast images preserved information relevant to Ki-67 prediction tasks. The proposed method achieved the best performance among compared methods. For whole-image evaluation, it yielded SSIM of 70.8, PSNR of 25.1, and NRMSE of 41.6, outperforming other baselines. In tumor-focused and tumor-plus-BPE evaluations, the proposed method also showed the strongest quantitative results, with SSIM up to 64.7 and PSNR up to 13.4 in the more challenging regional settings. For downstream classification of high versus low Ki-67, synthetic-image-based evaluation showed similar performance relative to real-image-based evaluation, without statistically significant differences across real-to-real, fake-to-fake, fake-to-real, and real-to-fake testing settings. An anatomy-aware image synthesis framework can generate post-contrast breast MRI from pre-contrast images with improved structural fidelity and enhanced lesion/BPE realism. Synthetic post-contrast images may retain clinically relevant information for selected downstream biomarker prediction tasks, supporting further investigation of contrast-free MRI workflows.
\end{abstract}
\clearpage

\endgroup
\section{Introduction}

Dynamic contrast-enhanced (DCE) breast MRI plays an important role in breast lesion characterization and treatment assessment, but gadolinium-based contrast administration adds cost, workflow complexity, and potential safety concerns \cite{ref3} \cite{ref5}. A computational approach that generates post-contrast images from pre-contrast scans could reduce reliance on administered contrast while preserving diagnostically relevant enhancement patterns.

Recent studies have explored this problem using different generative paradigms. Kim et al. proposed a tumor-attentive, segmentation-guided GAN that emphasized lesion-focused synthesis, showing the importance of explicitly modeling tumor regions during post-contrast image generation \cite{ref4}. Osuala et al. further demonstrated that pre- to post-contrast breast MRI synthesis can support downstream tumor segmentation, suggesting that synthesis quality should be assessed not only by visual fidelity but also by clinical utility \cite{ref6}. More recently, Osuala et al. extended this direction with a multi-condition latent diffusion framework to model contrast kinetics across temporal sequences, highlighting the potential of diffusion-based methods for simulating enhancement dynamics beyond single-sequence translation \cite{ref7}. In parallel, Zhou et al. examined how to quantify the quality of GAN-synthesized post-contrast breast MRI and showed that conventional similarity metrics alone may be insufficient to determine whether synthesized images preserve clinically meaningful enhancement information \cite{ref16}. Together, these studies suggest that both generation strategy and evaluation protocol are critical for reliable breast MRI synthesis.

Despite this progress, accurate synthesis of breast MRI remains challenging because diagnostically important information is concentrated in localized and anatomically meaningful regions, including tumor enhancement and background parenchymal enhancement (BPE) \cite{ref1}. Tumor focused methods improve attention to lesion regions, but they may not explicitly preserve broader enhancement-related structures such as BPE \cite{ref4}. Task-oriented synthesis methods demonstrate downstream value, but they do not necessarily enforce preservation of shared anatomy and enhancement patterns simultaneously. Diffusion-based approaches may better capture contrast kinetics, yet they also introduce greater modeling complexity and still require mechanisms to preserve subtle clinically relevant regions. Overall, prior work suggests that region-aware supervision is important, yet a unified framework that explicitly incorporates both anatomical structure and enhancement-specific priors remains underexplored. Reader-based evaluation is essential because the usefulness of synthesized breast MRI ultimately depends on whether image quality is acceptable for human interpretation, whereas a downstream AI modeling task is important because it assesses whether quantitative analysis of synthesized images preserve biologically relevant tumor information \cite{ref2}. Prior studies support the importance of each perspective individually, but they are still rarely examined together within a unified synthesis framework.

In this work, we propose an anatomy-aware framework for synthesizing post-contrast breast MRI from pre-contrast images. Built upon Pix2PixHD~\cite{ref11}, a high-resolution conditional GAN for image-to-image translation, our method incorporates additional supervision from entire breast region, lesion area, and BPE regions. Unlike conventional whole-image supervision, this design explicitly constrains both shared anatomical structure and enhancement related regions that are critical for breast imaging analysis and cancer diagnosis. We hypothesize that such anatomy-aware and region specific guidance can improve synthesis quality over standard image-to-image translation baselines. We further evaluate whether the synthesized images preserve clinically relevant information for a downstream AI modeling, Ki-67 prediction, thereby assessing not only visual realism but also functional utility. In addition, we conducted a blinded reader study involving two experienced breast radiologists to assess real versus synthesized MRI images. We developed a tailored user interface to minimize bias, ensure blinding, and automatically record results, while also facilitating efficient image loading and assessment entry for the radiologists.

\section{Materials and Methods}

\subsection{Study Cohorts and Datasets}

This retrospective study complied with the Health Insurance Portability and Accountability Act. The institutional review board approved the study and waived the requirement for written informed consent because anonymized data were used retrospectively. Breast MRI data from 649 patients diagnosed with breast cancer yielded 6,251 paired pre-contrast and post-contrast images from tumor-containing slices. These data were used to develop and evaluate an image synthesis model designed to generate post-contrast breast MRI from pre-contrast images using five-fold cross-validation at the patient level. Patients were divided into five nonoverlapping folds, with all slices from each patient assigned to the same fold. In each iteration, four folds were used for training and the remaining fold for testing, ensuring no patient overlap between the training and test sets. Detailed information is provided in Table~\ref{tab:1}.

\begin{table}[!htbp]
\centering
\caption{Clinicopathologic characteristics stratified by Ki-67 status. High Ki-67 was defined as > 20, and low Ki-67 was defined as \(\leq\)20.}
\label{tab:1}
\small\setlength{\tabcolsep}{4pt}\renewcommand{\arraystretch}{1.18}
\begin{tabularx}{\textwidth}{@{}>{\raggedright\arraybackslash}Xcccc@{}}
\toprule
Characteristic & Total & High Ki-67 & Low Ki-67 & $P$ value \\
 & ($n=649$) & ($n=215$) & ($n=431$) & \\
\midrule
Age (y) & 57.3 \(\pm\) 10.4 & 56.2 \(\pm\) 10.3 & 57.8 \(\pm\) 10.4 & 0.066 \\
Tumor size (cm) & 1.9 \(\pm\) 1.2 & 1.9 \(\pm\) 1.1 & 1.8 \(\pm\) 1.2 & 0.695 \\
Oncotype DX RS & 17.5 \(\pm\) 10.3 & 21.9 \(\pm\) 10.9 & 15.4 \(\pm\) 9.3 & < 0.001 \\
Tumor subtype &  &  &  & 0.001 \\
Invasive ductal carcinoma & 526 (81.0\%) & 193 (89.8\%) & 331 (76.8\%) &  \\
Invasive lobular carcinoma & 90 (13.9\%) & 15 (7.0\%) & 74 (17.2\%) &  \\
Invasive, mixed ductal and lobular carcinoma & 8 (1.2\%) & 3 (1.4\%) & 5 (1.2\%) &  \\
Invasive tubular carcinoma & 0 (0.0\%) & 0 (0.0\%) & 0 (0.0\%) &  \\
Invasive, mixed ductal and mucinous carcinoma & 5 (0.8\%) & 0 (0.0\%) & 5 (1.2\%) &  \\
Mucinous & 6 (0.9\%) & 3 (1.4\%) & 3 (0.7\%) &  \\
Invasive, mixed ductal and tubular carcinoma & 8 (1.2\%) & 0 (0.0\%) & 8 (1.9\%) &  \\
\bottomrule\end{tabularx}
\end{table}

\subsection{Breast DCE-MRI}

All women underwent imaging in the prone position on a 1.5-T scanner (SIGNA MRI family, MR software version 12LX, MR software release 12.0--15.0; GE HealthCare) using a dedicated seven channel surface-array breast coil. The imaging parameters included an in-plane resolution of 512 (Pixel) \(\times\) 512 (Pixel), a slice thickness of 2-3 mm, and a field of view of 28--34 cm.

\subsection{Model Architecture}

\begin{figure}[!htbp]
\centering
\includegraphics[width=\textwidth]{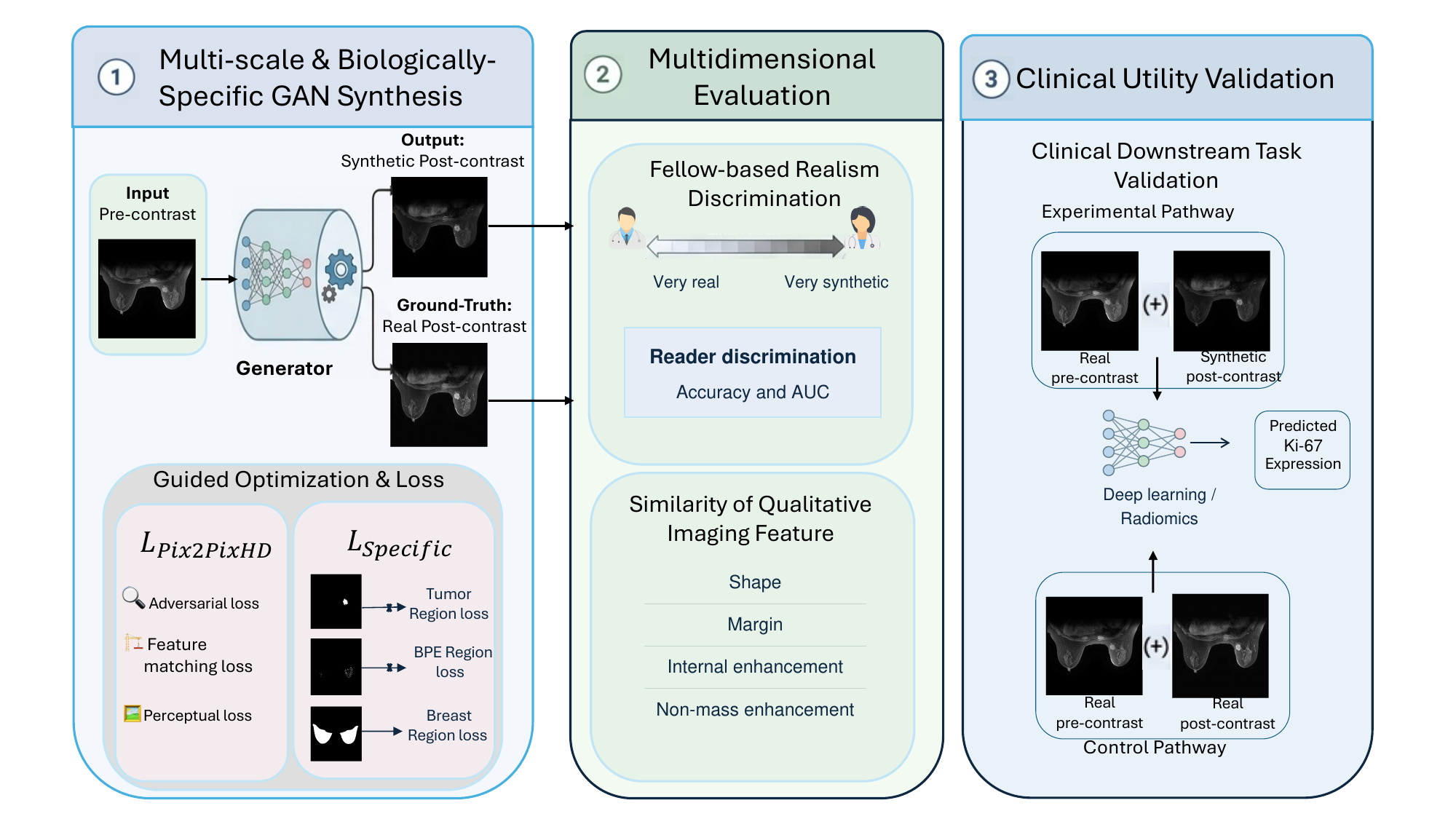}
\caption{Overview of the proposed biologically guided synthesis framework for generating postcontrast breast MRI from pre-contrast input. The model is optimized using a combination of adversarial, feature matching, perceptual, tumor-region, breast-region, and background parenchymal enhancement (BPE) region losses to improve both visual fidelity and biologically meaningful enhancement patterns. The framework is evaluated through a multidimensional strategy including radiologist-based realism assessment, similarity of qualitative imaging features, and downstream clinical utility validation.}
\label{fig:1}
\end{figure}

As shown in Figure~\ref{fig:1}, the proposed method was an anatomy-aware extension of Pix2PixHD \cite{ref11}, a high-resolution conditional image-to-image translation framework for synthesizing post-contrast breast MRI from pre-contrast inputs. We refer to the resulting model as Pix2PixH-BTC, a Pix2PixHD-based model that emphasizes BPE preservation, tumor-region fidelity, and breast-region consistency. Specifically, the generator received the real pre-contrast image and produced a synthetic post-contrast image, while the discriminator network was designed at both full resolution and half resolution to evaluate realism at complementary spatial scales. The full-resolution discriminator encouraged preservation of fine local structures and detailed enhancement patterns, whereas the half-resolution discriminator emphasized global anatomical layout and coarse-scale consistency. During training, breast masks, lesion regions, and BPE regions were used as anatomy-aware supervision to guide the model toward more realistic and clinically relevant enhancement patterns. During testing and image generation, however, the model required only the pre-contrast image as input and did not require breast masks, lesion annotations, or BPE-region information.

To further constrain the synthesis, the framework incorporated several auxiliary supervision terms beyond the standard adversarial objective. A perceptual feature loss, computed using a pretrained VGG network \cite{ref8}, encouraged the synthesized image to match the real post-contrast image in higher-level feature space rather than only at the pixel level, thereby promoting perceptual realism and structural similarity. A lesion-region supervision loss was introduced to focus the model on tumor-containing regions, ensuring that diagnostically important lesion enhancement patterns were preserved during synthesis. In parallel, a BPE-region supervision loss specifically constrained the background parenchymal enhancement regions, which are clinically meaningful yet often subtle and difficult to reproduce with whole-image supervision alone.

In addition, the model included a shared-region consistency term derived from breast masks. This term regularized intensity changes within the anatomically shared breast region, encouraging the model to generate spatially coherent enhancement while avoiding unrealistic fluctuations or artifacts outside the expected tissue boundaries. By explicitly supervising breast anatomy, lesion regions, and BPE regions, the framework moved beyond conventional whole-image translation and better aligned the synthesized output with clinically relevant breast MRI structures.

Overall, the final loss function combined adversarial loss, perceptual loss, lesion-region loss, BPE-region loss, and shared-region consistency loss. Together, these components were intended to enforce several complementary properties: adversarial learning promoted overall visual realism; the multi-scale discriminators captured both global structure and fine detail; perceptual supervision improved feature-level similarity; lesion and BPE losses enhanced fidelity in clinically important subregions; and the shared-region consistency term preserved anatomically plausible and spatially coherent enhancement patterns within the breast. This design allowed the model to synthesize postcontrast images that were not only visually realistic, but also better aligned with the anatomical and functional characteristics required for downstream clinical analysis.

\subsection{Automated BPE Quantification}

\begin{figure}[!htbp]
\centering
\includegraphics[width=\textwidth]{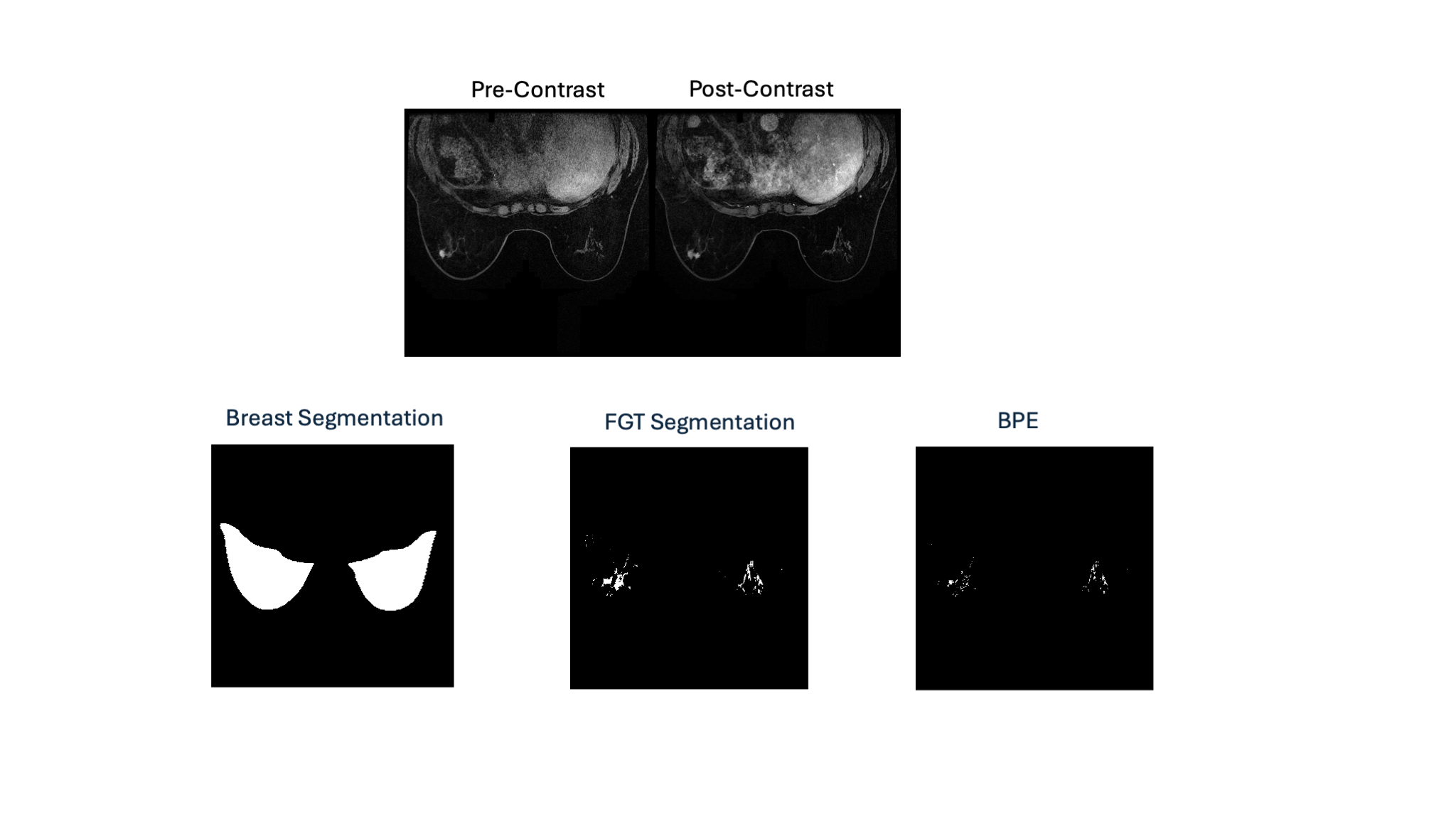}
\caption{Representative middle-slice breast MRI images and segmentation-based biomarker extraction pipeline. Representative middle-slice images from our breast MRI dataset are shown for the pre-contrast image, real post-contrast image. Breast segmentation and fibroglandular tissue (FGT) segmentation were applied to extract anatomically relevant regions for downstream biomarker analysis. BPE-related region information was used during model training as anatomy-aware supervision to guide synthesis of biologically meaningful enhancement patterns, rather than as an additional input during testing. The extracted breast and FGT regions provide visual context for assessing whether the synthesized post-contrast images preserve clinically relevant tissue structures and enhancement characteristics beyond conventional pixel-wise synthesis metrics.}
\label{fig:3}
\end{figure}

Previously validated computer algorithms were applied to automatically quantify the absolute volume of BPE and its relative amount over the whole-breast volume \cite{ref1,ref12,ref13}. First, a pretrained U-Net segmented the breast region from the background \cite{ref14,ref15}. Next, fibroglandular tissue (FGT) was segmented by normalizing pixel intensities within the breast region and applying fuzzy c-means clustering into three clusters; the brightest cluster was selected as FGT. Finally, tumor regions were excluded, pre-contrast and post-contrast differences were computed, and thresholding was applied to isolate clinically relevant enhancement areas corresponding to BPE. Representative examples of breast segmentation, FGT segmentation, and BPE masks were shown for multiple cases. Details are shown in Figure~\ref{fig:3}.

\subsection{Quantitative Evaluation}

Image quality was assessed using SSIM, PSNR, and NRMSE. Evaluations were reported for whole-image synthesis and for more focused settings involving tumor areas and tumor-plus-BPE regions. Representative quantitative comparisons were also shown between ground-truth postcontrast images, GAN outputs, Pix2PixHD outputs, and the proposed method across multiple cases.

\subsection{Reader Study}

To assess the perceptual realism and qualitative imaging fidelity of the synthesized post-contrast breast MRI images, a reader study was performed using a dedicated interactive software tool.

\begin{figure}[!htbp]
\centering
\includegraphics[width=\textwidth]{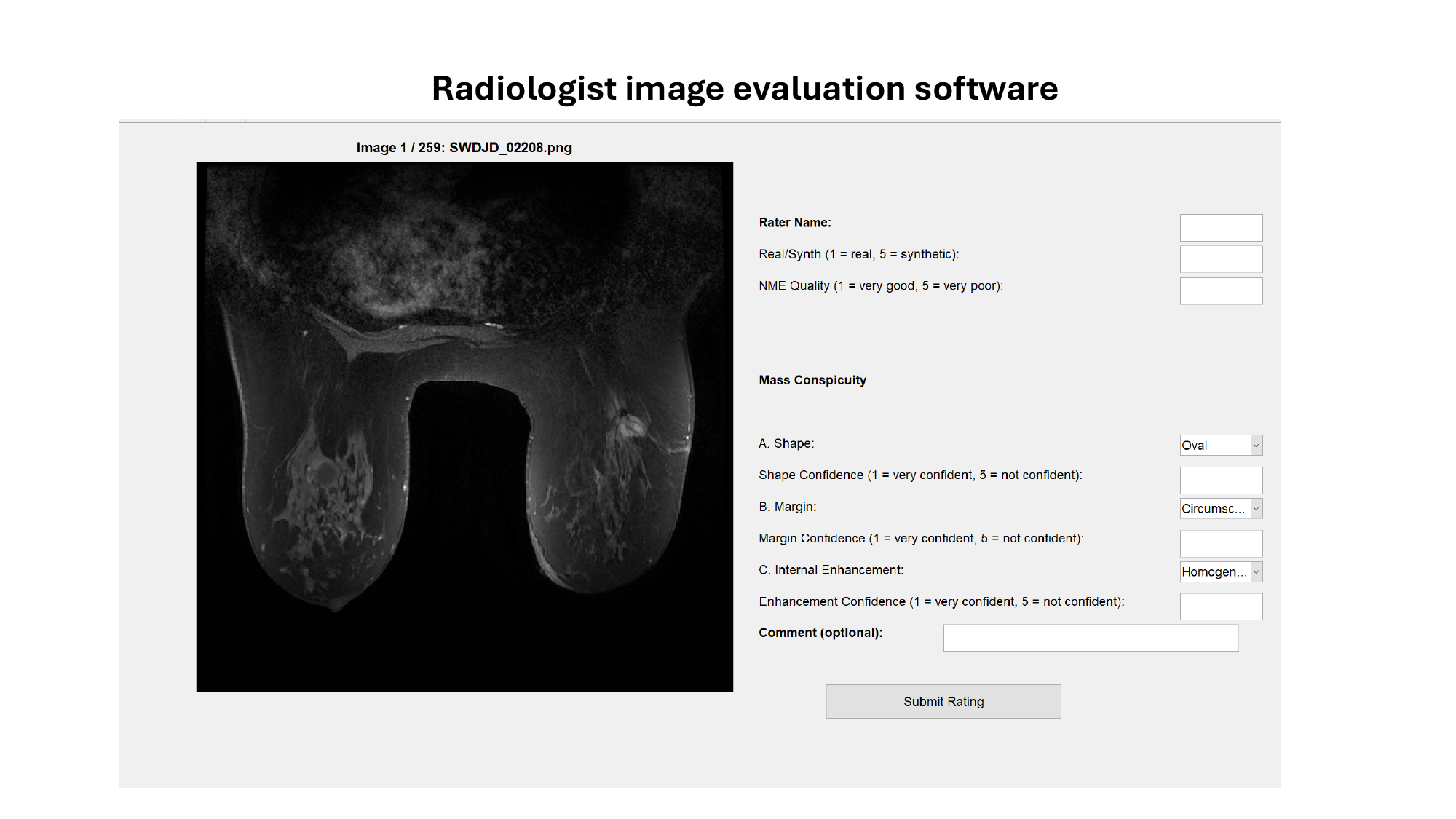}
\caption{Reader study interface and results for radiologist-based realism evaluation of synthesized post-contrast breast MRI. Radiologists or fellows scored generated images on a scale ranging from very real to very synthetic to assess perceptual realism.}
\label{fig:2}
\end{figure}

The software randomly displayed real and synthesized images and allowed readers to record their assessments in a standardized interface as shown in Figure~\ref{fig:2}. The platform was designed for online use, enabling remote access while minimizing data leakage and potential reading bias.

The reader study included 130 MRI scans from patients with biopsy-proven breast cancer. For each case, 1 real first post-contrast MRI image and 1 synthesized first post-contrast image generated from the corresponding pre-contrast scan were included, yielding 130 real and 130 synthesized images. The evaluated image corresponded to the middle tumor slice.

Two breast radiology fellows independently reviewed the images. For each case, they assessed key qualitative imaging features relevant to breast cancer diagnosis, including lesion shape, lesion margin, internal enhancement characteristics, and the quality of non-mass enhancement (NME). Each assessment was scored on a 5-point scale, where 1 indicated ``very confident'' and 5 indicated ``not confident''. In addition, they assigned a score (1--5) to indicate how synthetic each image appeared, where 1 represented ``very real'' and 5 represented ``very synthetic,'' based on their assessment.

Reader discrimination performance for distinguishing real from synthesized images was assessed using the area under the receiver operating characteristic curve (AUC) and classification accuracy. To evaluate the similarity of qualitative imaging features between real and synthesized images, categorical variables were compared using the chi-square test and summarized with Cramer's V effect size, whereas continuous variables were compared using the two-sample Kolmogorov Smirnov test and summarized with Cohen's d. Similarity was interpreted as very high for effect sizes of 0.0--0.2, high for 0.2--0.5, moderate for 0.5--0.8, and low for greater than 0.8.

\subsection{Downstream Biomarker Prediction}

The study also examined whether synthetic post-contrast images preserved features useful for downstream prediction of Ki-67. Both radiomics-based \cite{ref10} and deep learning workflows were explored. For Ki-67 classification, a threshold of 20 was used to define high- and low-expression groups. After excluding cases with missing Ki-67 values, 648 cases were included, consisting of 216 high Ki-67 cases and 432 low Ki-67 cases. Five-fold cross-validation was performed, and statistical comparisons were conducted using independent two-sample t-tests.

For the deep learning experiments, a pretrained EfficientNet-B2 \cite{ref9} was used and trained for 15 epochs with an input size of 128 \(\times\) 128. Inputs included whole images, tumor-masked regions, and cropped tumor areas. For the cropped-area setting, the tumor region was extracted based on the lesion mask, with an 8-pixel margin added around the mask to preserve adjacent contextual tissue. Additional experiments compared real-to-real, fake-to-fake, fake-to-real, and real-to-fake settings to assess whether synthetic images preserved Ki67-relevant imaging information across within-domain and cross-domain evaluation scenarios.

\section{Results}

\subsection{Quantitative Synthesis Performance}

As shown in Table~\ref{tab:2}, the proposed method achieved the best quantitative performance among all compared models. In whole-image evaluation, the proposed method attained SSIM of 70.8, PSNR of 25.1, and NRMSE of 41.6, compared with 70.1, 24.8, and 43.5 for Pix2PixHD-BTM and 69.3, 24.5, and 45.3 for Pix2PixHD. Diffusion and Pix2Pix showed lower performance.

In the tumor-area-focused evaluation, the proposed method again performed best, with SSIM of 64.4, PSNR of 13.2, and NRMSE of 38.8. In the tumor-plus-BPE setting, it achieved SSIM of 64.7, PSNR of 13.4, and NRMSE of 38.9, outperforming all baselines. These findings suggest that the anatomy-aware constraints improved not only whole-image realism but also fidelity in clinically meaningful enhancement regions.

\begin{table}[!htbp]
\centering
\caption{Quantitative image quality comparison of synthesis methods. Results are reported for whole-image, tumor-area, and tumor-area + background parenchymal enhancement (BPE) mask evaluations. Values in brackets indicate confidence intervals. Higher SSIM and PSNR indicate better image quality, whereas lower NRMSE indicates better performance.}
\label{tab:2}
\small\setlength{\tabcolsep}{5pt}\renewcommand{\arraystretch}{1.12}
\begin{tabularx}{\textwidth}{@{}>{\raggedright\arraybackslash}Xccc@{}}
\toprule
\multicolumn{4}{@{}l}{\textbf{Whole Image}} \\
Method & SSIM & PSNR & NRMSE \\
\midrule
Pix2Pix & 67.8 [67.5, 68.1] & 23.4 [23.1, 23.5] & 52.0 [49.8, 52.2] \\
Pix2PixHD & 69.3 [69.0, 69.6] & 24.5 [24.2, 24.7] & 45.3 [44.9, 45.7] \\
Diffusion & 67.9 [67.3, 68.5] & 23.6 [23.2, 23.9] & 44.3 [44.0, 44.5] \\
Pix2PixHD-TM & 69.9 [69.3, 70.5] & 24.6 [24.3, 24.9] & 43.9 [43.6, 44.3] \\
Pix2PixHD-BTM & 70.1 [69.6, 71.0] & 24.8 [24.4, 25.2] & 43.5 [43.3, 43.8] \\
\textbf{Pix2PixH-BTC (Proposed)} & \textbf{70.8 [69.8, 71.8]} & \textbf{25.1 [24.9, 25.4]} & \textbf{41.6 [41.1, 42.1]} \\
\midrule
\multicolumn{4}{@{}l}{\textbf{Tumor Area}} \\
Method & SSIM & PSNR & NRMSE \\
\midrule
Pix2Pix & 58.7 [55.6, 61.8] & 12.1 [11.5, 12.8] & 44.9 [43.9, 45.9] \\
Pix2PixHD & 60.2 [56.8, 63.6] & 12.2 [11.8, 12.6] & 42.1 [41.7, 42.5] \\
Diffusion & 59.9 [55.9, 63.9] & 11.6 [11.2, 12.0] & 43.3 [42.8, 43.8] \\
Pix2PixHD-TM & 61.4 [59.1, 63.6] & 12.7 [12.0, 13.4] & 41.9 [41.4, 42.4] \\
Pix2PixHD-BTM & 61.1 [58.6, 63.6] & 12.8 [12.2, 13.4] & 41.7 [41.3, 42.1] \\
\textbf{Pix2PixH-BTC (Proposed)} & \textbf{64.4 [60.2, 68.6]} & \textbf{13.2 [12.9, 13.5]} & \textbf{38.8 [38.2, 39.4]} \\
\midrule
\multicolumn{4}{@{}l}{\textbf{Tumor Area + BPE Mask}} \\
Method & SSIM & PSNR & NRMSE \\
\midrule
Pix2Pix & 58.8 [55.8, 61.8] & 11.9 [11.2, 12.6] & 45.1 [44.6, 45.6] \\
Pix2PixHD & 60.1 [57.7, 62.5] & 12.3 [12.0, 12.6] & 42.3 [42.0, 42.6] \\
Diffusion & 59.7 [55.7, 63.7] & 11.7 [11.4, 12.0] & 43.5 [42.9, 44.1] \\
Pix2PixHD-TM & 61.4 [59.4, 63.4] & 12.4 [11.8, 13.0] & 42.3 [41.8, 42.8] \\
Pix2PixHD-BTM & 61.7 [59.0, 64.4] & 12.9 [12.4, 13.4] & 42.1 [41.7, 42.4] \\
\textbf{Pix2PixH-BTC (Proposed)} & \textbf{64.7 [60.2, 68.9]} & \textbf{13.4 [13.1, 13.7]} & \textbf{38.9 [38.4, 39.5]} \\
\bottomrule\end{tabularx}
\end{table}

\subsection{Qualitative Findings}

\begin{figure}[p]
\centering
\includegraphics[width=\linewidth,height=\dimexpr\textheight-1.5in\relax,keepaspectratio]{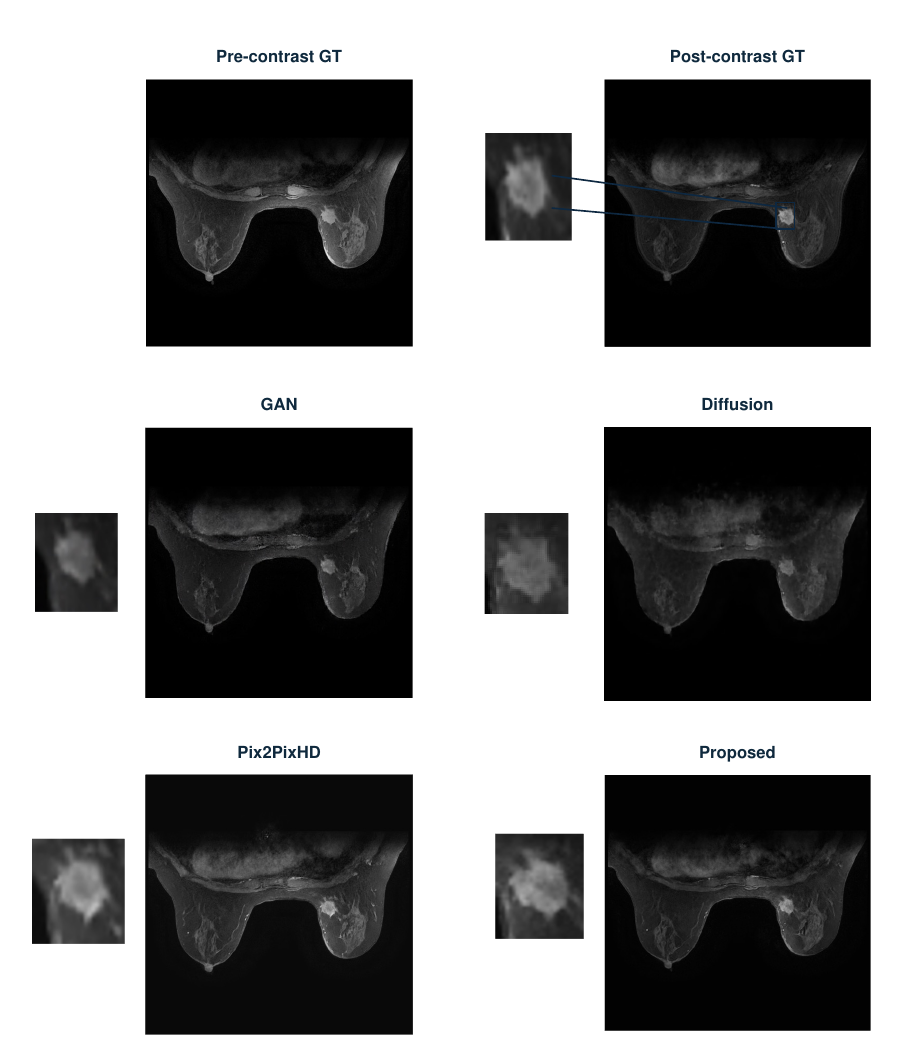}
\caption{Qualitative comparison of synthesized post-contrast breast MRI generated by different methods. Representative examples are shown for GAN, Pix2PixHD, diffusion, and the proposed method, alongside the real pre-contrast input and ground-truth post-contrast image. Compared with competing approaches, the proposed method produces more realistic enhancement patterns and better preserves anatomical structure and lesion-related appearance.}
\label{fig:4}
\end{figure}

\begin{figure}[p]
\centering
\includegraphics[width=\linewidth,height=\dimexpr\textheight-1.5in\relax,keepaspectratio]{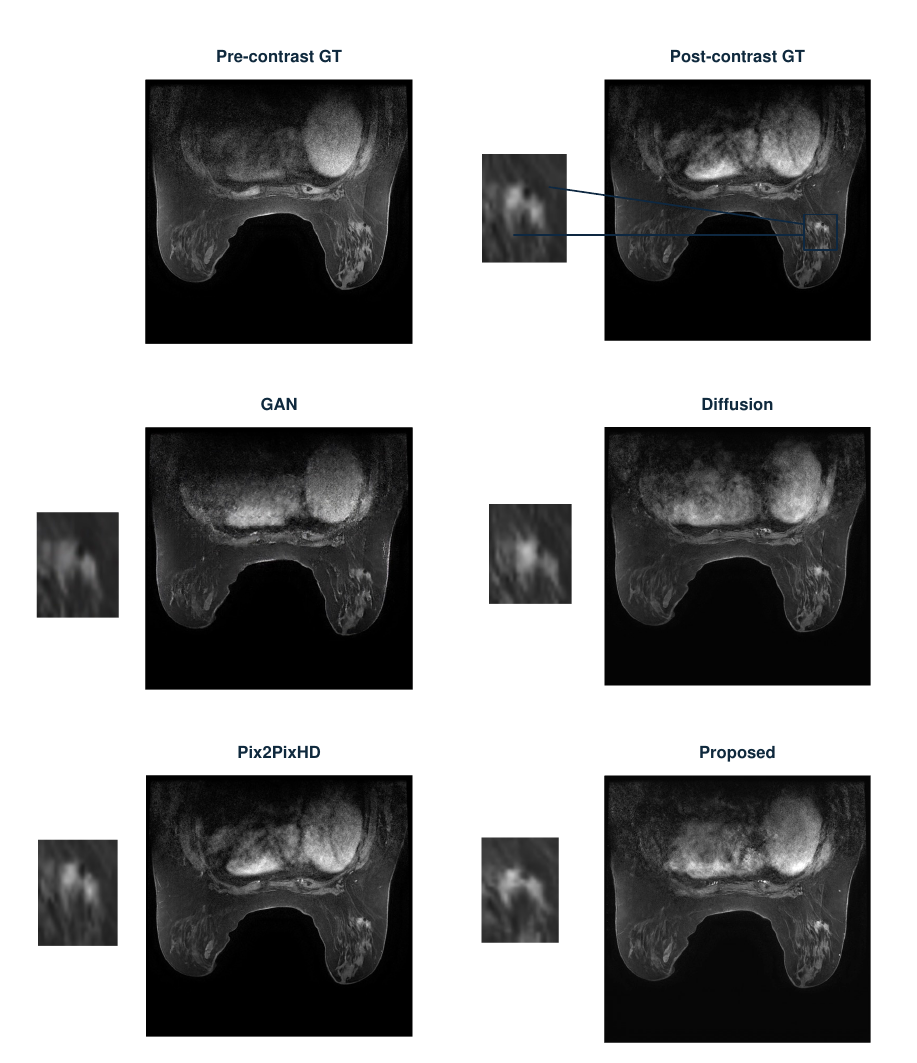}
\caption{Qualitative comparison of synthesized post-contrast breast MRI in a second representative case. GAN, diffusion, Pix2PixHD, and the proposed method are shown alongside the pre-contrast input and ground-truth post-contrast image. Magnified lesion insets highlight local enhancement appearance across the compared methods.}
\label{fig:5}
\end{figure}

Representative qualitative results demonstrated that the proposed method more closely reproduced lesion conspicuity and enhancement appearance than the compared GAN and Pix2PixHD outputs. Across several visualized cases, the synthetic images generated by the proposed method appeared to preserve more realistic local enhancement patterns in the highlighted lesion regions while maintaining global breast anatomy. The qualitative findings consistently showed closer visual agreement between the proposed method and ground-truth post-contrast images than with the baseline approaches as shown in Figures~\ref{fig:4} and~\ref{fig:5}.

\clearpage
\clearpage

\clearpage
\clearpage

\subsection{Radiologist Assessment}

In the reader study, 2 breast radiology fellows evaluated 260 images, including 130 real first post-contrast breast MRI images and 130 synthesized first post-contrast images.

Qualitative imaging feature analysis showed high to very high similarity between real and synthesized images for both lesion and contrast enhancement descriptors. For lesion shape, there was no significant difference between real and synthesized images using the chi-square test for categorical variables (P = 0.468), with a Cramer's V of 0.076, indicating very high similarity. Similarly, lesion margin showed no significant difference (P = 0.806), with a Cramer's V of 0.0407, also indicating very high similarity. Lesion internal enhancement characteristics were also comparable between real and synthesized images (P = 0.7485; Cramer's V = 0.0685).

For non-mass enhancement (NME) quality (a continuous variable), the difference between real and synthesized images was statistically significant by the two-sample Kolmogorov-Smirnov test (P = 0.0010); however, the effect size was modest (Cohen's d = -0.3967), corresponding to high similarity rather than a large discrepancy. Overall, these results suggest that the synthesized post-contrast images preserved several important qualitative imaging characteristics, although subtle perceptual differences remained detectable.

Despite this, readers were able to distinguish real from synthesized images with moderate performance. As shown in Table~\ref{tab:3}, Reader 1 achieved an AUC of 0.787 and an accuracy of 0.769, while Reader 2 achieved an AUC of 0.775 and an accuracy of 0.731. The overall mean performance was an AUC of 0.779 and an accuracy of 0.735. These findings indicate that although the synthesized images were visually realistic in terms of key imaging features, radiologists could still identify synthetic images better than chance.

\begin{table}[!htbp]
\centering
\caption{Radiologist reader study results and similarity of qualitative imaging features between real and synthesized post-contrast breast MRI images. Reader performance was evaluated using area under the receiver operating characteristic curve (AUC) and accuracy. Similarity between qualitative imaging features extracted from real and synthesized images was assessed using chi-square tests with Cramer's V for categorical variables and the two-sample Kolmogorov-- Smirnov test with Cohen's d for the continuous variable. Smaller effect sizes indicate greater similarity.}
\label{tab:3}
\small\setlength{\tabcolsep}{4pt}\renewcommand{\arraystretch}{1.2}
\begin{tabularx}{\textwidth}{@{}>{\raggedright\arraybackslash}Xll@{}}
\toprule
\multicolumn{3}{@{}l}{\textbf{Reader discrimination}} \\
Reader & AUC & Accuracy \\ \midrule
Reader 1 & 0.787 & 0.769 \\
Reader 2 & 0.775 & 0.731 \\
Overall & 0.779 & 0.735 \\
\midrule\multicolumn{3}{@{}l}{\textbf{Feature similarity}} \\
Variable & $P$ value & Effect size \\ \midrule
Shape & 0.4685 & Cramer's $V=0.0764$ \\
Margin & 0.8059 & Cramer's $V=0.0407$ \\
Internal enhancement & 0.7485 & Cramer's $V=0.0685$ \\
NME quality & 0.0010 & Cohen's $d=-0.3967$ \\
\bottomrule\end{tabularx}
\end{table}

\subsection{Downstream task for Ki-67 Prediction}

For Ki-67 classification, as shown in Table~\ref{tab:4}, EfficientNet-B2 achieved a mean AUC of 0.602 when trained and evaluated using acquired post-contrast images. The mean AUCs were 0.600 when trained and evaluated using synthetic images, 0.598 when trained using synthetic images and evaluated using acquired images, and 0.595 when trained using acquired images and evaluated using synthetic images. The corresponding mean AUCs for the radiomics-based model were 0.599, 0.597, 0.582, and 0.591, respectively. For both models, none of the configurations involving synthetic images differed significantly from the configuration trained and evaluated using acquired post-contrast images (all P \(\geq\) .46).

\begin{table}[!htbp]
\centering
\caption{Comparison of performance under different train--test combinations of ground-truth (GT) and synthetic datasets for Efficient-B2 and radiomic features. GT\(\rightarrow\)GT is treated as the reference setting within each model. Results are reported as mean \(\pm\) standard deviation across folds. NS means No Significant Difference}
\label{tab:4}
\small\setlength{\tabcolsep}{4pt}\renewcommand{\arraystretch}{1.2}
\begin{tabularx}{\textwidth}{@{}>{\raggedright\arraybackslash}Xccc@{}}
\toprule
\multicolumn{4}{@{}l}{\textbf{Efficient-B2}} \\
Setting & Performance & $P$ value & Significance \\ \midrule
GT \(\rightarrow\) GT & 0.602\(\pm\)0.021 & -- & Reference \\
Synthetic \(\rightarrow\) Synthetic & 0.600\(\pm\)0.030 & 0.90 & NS \\
Synthetic \(\rightarrow\) GT & 0.598\(\pm\)0.021 & 0.80 & NS \\
GT \(\rightarrow\) Synthetic & 0.595\(\pm\)0.032 & 0.68 & NS \\
\midrule
\multicolumn{4}{@{}l}{\textbf{Radiomic features}} \\
Setting & Performance & $P$ value & Significance \\ \midrule
GT \(\rightarrow\) GT & 0.599\(\pm\)0.039 & -- & Reference \\
Synthetic \(\rightarrow\) Synthetic & 0.597\(\pm\)0.022 & 0.93 & NS \\
Synthetic \(\rightarrow\) GT & 0.582\(\pm\)0.021 & 0.46 & NS \\
GT \(\rightarrow\) Synthetic & 0.591\(\pm\)0.043 & 0.71 & NS \\
\bottomrule\end{tabularx}
\end{table}

\section{Discussion}

This study presents an anatomy-aware framework for generating post-contrast breast MRI from pre-contrast images. By augmenting a Pix2PixHD backbone with breast mask consistency and localized supervision over lesion and BPE regions, the proposed method improved quantitative synthesis performance over conventional GAN, Pix2PixHD, and diffusion-based baselines. The improvements were observed not only on whole-image metrics but also in targeted evaluations focused on lesion and enhancement regions, which are particularly important for breast MRI interpretation.

A key contribution of the framework is its explicit incorporation of breast anatomy and enhancement specific priors. Breast MRI synthesis is challenging because diagnostically relevant contrast changes are spatially sparse and anatomically constrained. Standard image-level losses may favor global realism while underemphasizing subtle yet clinically meaningful enhancement patterns. The proposed approach addressed this issue by combining region-specific supervision for lesion and BPE areas with a shared-region consistency term, encouraging coherent enhancement behavior within anatomical structures. The qualitative examples support this interpretation, showing better preservation of focal lesion appearance relative to baseline methods.

The reader study showed that radiologist-extracted qualitative imaging features related to lesion characteristics and contrast enhancement did not differ significantly between synthetic and real post-contrast MRI images, demonstrating high similarity. These findings are promising and suggest that key imaging features relevant for the specific Ki-67 analysis are preserved in the images generated by our model. Meanwhile, we noticed that readers were still able to distinguish synthetic from real images better than chance, indicating that the perceptible differences remain between generated and acquired post-contrast MR images. This could be due to multiple factors such as artifacts introduced during image synthesis. Artifacts outside the breast regions may reveal a synthetic image but they will not directly affect analysis on breast or lesion areas. Future work will need to further investigate these differences and improve the model accordingly.

Downstream experiments indicated that synthetic post-contrast images preserve quantitative imaging information relevant to the Ki-67-related classification tasks, as no significant differences were observed across several real/synthetic train-test configurations. Such an evaluation method has meaningful clinical values in addition to pure technical evaluation.

This study has several limitations. First, we used breast DCE-MRI data from a single scanner at a single institution for model development and evaluation in this proof-of-concept study. This was intentional to maintain data homogeneity and enable the model to focus on learning contrast enhancement patterns rather than variability across scanners. Future work will involve models that incorporate data from multiple scanners to improve generalizability. Second, our model used only pre-contrast MRI as input, while previous studies have examined the use of diffusion-weighted imaging and/or other MR sequences \cite{ref17}. Third, we did not systematically analyze the sources of radiologist disagreement, including whether these arose from artifacts in non-mass regions or differences in lesion appearance. Finally, downstream evaluation was limited to Ki-67 classification, and the predictive models were not specifically optimized to maximize performance on this task. Additional downstream tasks and task-specific optimization will contribute to further assessment of the synthetic images and our method.

In conclusion, anatomy-aware synthesis of post-contrast breast MRI from pre-contrast images is feasible and improves fidelity in clinically relevant regions. These findings support continued investigation of synthetic contrast enhancement as a step toward reducing reliance on contrast administration in selected breast MRI applications.

\clearpage
\section{ACKNOWLEDGMENTS}

This work was supported in part by a NIH Other Transaction research contract \#1OT2OD037972-01 and \#3OT2OD037972-01S1, a Radiological Society of North America (RSNA) Research \& Education Foundation Research Scholar Grant (Grant No. RSCH25-338; PI: Dooman Arefan), an Amazon Machine Learning Research Award, a PA Breast Cancer Coalition grant, a Jewish Healthcare Foundation grant, and the University of Pittsburgh Momentum Funds (Scaling Grant) for the Pittsburgh Center for AI Innovation in Medical Imaging. This work used Bridges-2 at the Pittsburgh Supercomputing Center through allocation [MED200006] from the Advanced Cyberinfrastructure Coordination Ecosystem: Services \& Support (ACCESS) program, which is supported by NSF grants \#2138259, \#2138286, \#2138307, \#2137603, and \#2138296. This research was also supported in computing resources by the University of Pittsburgh Center for Research Computing and Data (RRID: SCR\_022735) through the resources provided by the H2P cluster, which is supported by NSF award OAC-2117681. The views and conclusions contained in this document are those of the authors and should not be interpreted as representing official policies, either expressed or implied, of the NIH or NSF.

\end{document}